\documentclass[letterpaper]{article} 
\usepackage{aaai25}  
\usepackage{times}  
\usepackage{helvet}  
\usepackage{courier}  
\usepackage{booktabs}
\usepackage[hyphens]{url}  
\usepackage{graphicx} 
\usepackage{amsmath, amssymb}  
\usepackage{amsthm}

\usepackage{bm} 
\usepackage{natbib}  
\usepackage{caption} 
\usepackage{amsmath} 
\usepackage{algorithm}
\usepackage{algorithmic}
\usepackage{amssymb}
\usepackage{pifont}
\usepackage{array}

\usepackage[utf8]{inputenc} 
\usepackage[T1]{fontenc}    
\usepackage{url}            
\usepackage{booktabs}       
\usepackage{amsfonts}       
\usepackage{nicefrac}       
\usepackage{microtype}      
\usepackage{xcolor}         

\usepackage[accsupp]{axessibility} 
\usepackage{graphicx}
\usepackage{amsmath}
\usepackage{amssymb}
\usepackage{booktabs}
\usepackage{multirow}
\usepackage{colortbl}
\usepackage{adjustbox}
\usepackage{dsfont}

\usepackage{caption}
\usepackage{subfigure}
\usepackage{pifont}

\usepackage[ruled,vlined,linesnumbered,algo2e]{algorithm2e}
\usepackage{xcolor}

\newcommand{\uar}{\(\uparrow\)}

\newcommand{\tsb}{\textsubscript}

\newcommand{\appendixhead}%
{\begin{center}\textbf{{\large Appendix}}\end{center}
}

\usepackage{newfloat}
\usepackage{listings}
\usepackage{multirow}
\DeclareCaptionStyle{ruled}{labelfont=normalfont,labelsep=colon,strut=off} 
\floatstyle{ruled}
\newfloat{listing}{tb}{lst}{}
\floatname{listing}{Listing}
\title{
Harnessing Textual Semantic Priors for Knowledge Transfer and Refinement in CLIP-Driven Continual Learning}

\author {
    Lingfeng He\textsuperscript{\rm 1},
    De Cheng\textsuperscript{\rm 1, 2}\thanks{Corresponding author},
    Di Xu\textsuperscript{\rm 3},
    Huaijie Wang\textsuperscript{\rm 1},
    Nannan Wang\textsuperscript{\rm 1}
}
\affiliations {
    \textsuperscript{\rm 1}Xidian University\\
    \textsuperscript{\rm 1}
    National Engineering Laboratory for Integrated Aero-Space-Ground- Ocean Big Data Application Technology, China \\
    \textsuperscript{\rm 3}Huawei Technologies Co. Ltd\\
lingfenghe077@gmail.com,
dcheng@xidian.edu.cn,
xudi21@huawei.com,
huaijie\_wang@stu.xidian.edu.cn,
nnwang@xidian.edu.cn
}

\usepackage{bibentry}

\begin{document}

\maketitle




\begin{abstract}
Continual learning (CL) aims to equip models with the ability to learn from a stream of tasks without forgetting previous knowledge. With the progress of vision-language models like Contrastive Language-Image Pre-training (CLIP), their promise for CL has attracted increasing attention due to their strong generalizability. However, the potential of rich textual semantic priors in CLIP in addressing the stability–plasticity dilemma remains underexplored. During backbone training, most approaches transfer past knowledge without considering semantic relevance, leading to interference from unrelated tasks that disrupt the balance between stability and plasticity. Besides, while text-based classifiers provide strong generalization, they suffer from limited plasticity due to the inherent modality gap in CLIP. Visual classifiers help bridge this gap, but their prototypes lack rich and precise semantics. To address these challenges, we propose Semantic-Enriched Continual Adaptation (SECA), a unified framework that harnesses the anti-forgetting and structured nature of textual priors to guide semantic-aware knowledge transfer in the backbone and reinforce the semantic structure of the visual classifier. Specifically, a Semantic-Guided Adaptive Knowledge Transfer (SG-AKT) module is proposed to assess new images' relevance to diverse historical visual knowledge via textual cues, and aggregate relevant knowledge in an instance-adaptive manner as distillation signals. Moreover, a Semantic-Enhanced Visual Prototype Refinement (SE-VPR) module is introduced to refine visual prototypes using inter-class semantic relations captured in class-wise textual embeddings. Extensive experiments on multiple benchmarks validate the effectiveness of our approach. Our code is available at: \url{https://github.com/HHHLF/SECA_master}.
\end{abstract}

\section{Introduction}

\begin{figure}[!t]
\centering
\includegraphics[width=0.48\textwidth]{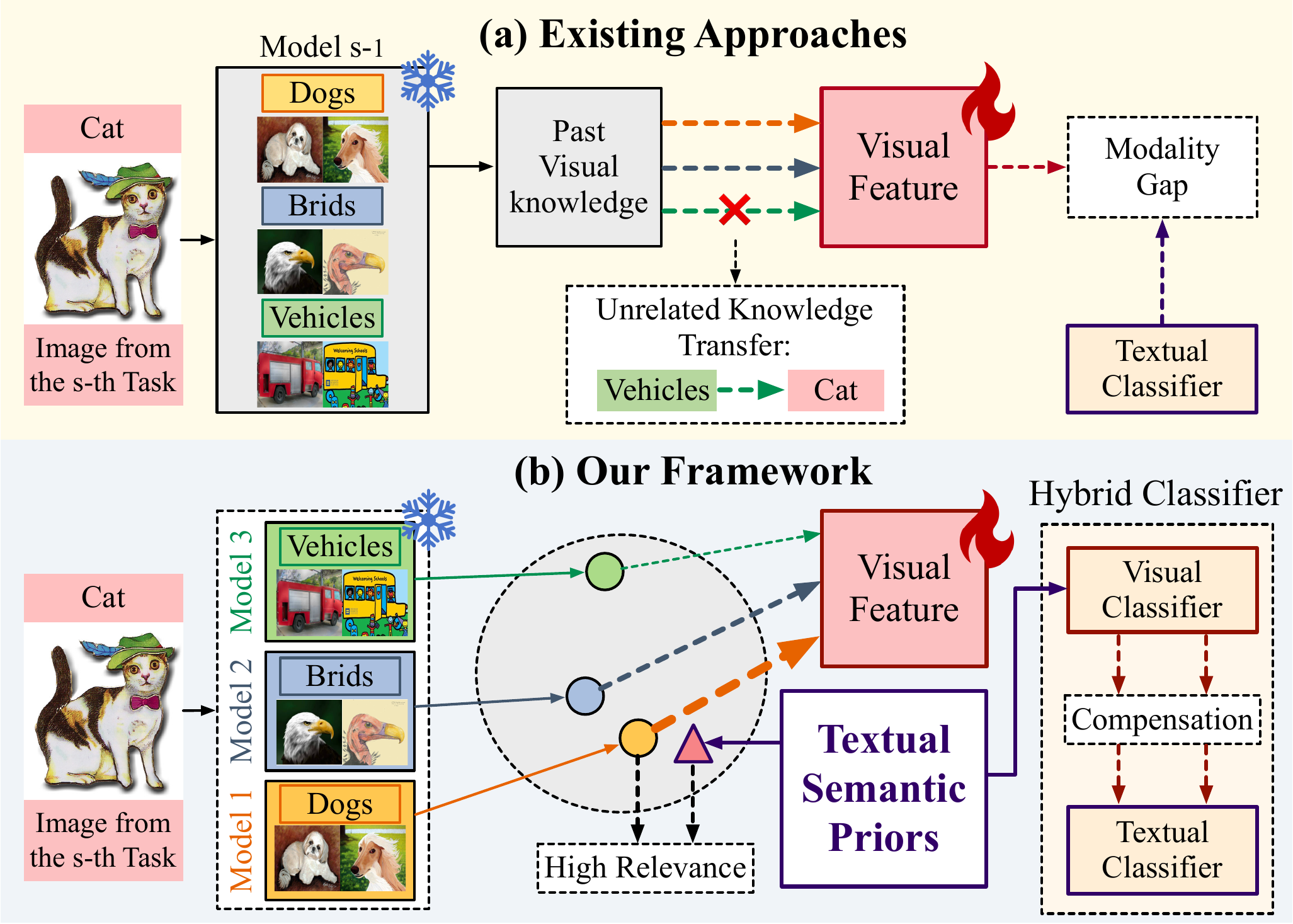}
\vspace{-5.0mm}
\caption{{
(a) Existing methods exhibit limited stability-plasticity trade-offs due to the unrelated knowledge transfer in backbone and the modality gap in classifier.
(b) Our SECA leverage textual priors to (1) prioritize transferring  relevant knowledge (Dog classes) when learning new classes (Cat), 
and (2) inject semantic relations into the visual classifier to bridge the modality gap, improving this trade-off.
}
}\label{fig: intro}
\vspace{-6.0mm}
\end{figure}

Continual learning (CL) aims to equip models with the ability to learn continuously from a stream of tasks~\cite{vptnspNIPS-24}, a key requirement for adapting to evolving open-world scenarios.
A fundamental challenge is \emph{the stability-plasticity dilemma}~\cite{kim2023stability, chen2023stability}, which requires models to preserve previously learned knowledge (stability) while adapting to new information (plasticity).
We investigate the challenging rehearsal-free Class-Incremental Learning (CIL) setting, where the model must continually learn new classes without rehearsal and predict over all seen categories without task identities.

Recent advances in vision-language models, such as Contrastive Language-Image Pre-training
(CLIP)~\cite{2021clip}, provide strong zero-shot capabilities and serve as promising backbones for CL. 
While several methods~\cite{PROOF-TPAMI25, rapfECCV24, clap4clip-NIPS24} 
leverage CLIP mainly as a powerful backbone, 
the potential of its textual semantic priors for improving the stability–plasticity trade-off remains underexplored.
To retain old knowledge in backbones, most approaches~\cite{EwC, AFC-CVPR22, PSRD-ICML23} rely on regularization or distillation to enforce consistency with model after the most recent task. 
However, such non-selective knowledge retention can lead to semantic interference, where unrelated or noisy knowledge from previous tasks hampers adaptation to new tasks and distorts earlier feature spaces.
As shown in Fig.~\ref{fig: intro}(a), when learning a new class like 'cat', prior knowledge from similar classes such as 'dog' or 'bird' can be helpful, while unrelated classes like 'vehicle' introduce interference that degrades learning.
As for classifiers, although
the text-only designs~\cite{rapfECCV24, clap4clip-NIPS24} offer strong generalization, they limit plasticity due to the inherent modality gap in CLIP.
Prior studies~\cite{liang2022mind, fahim2024s-its-not-a} attributes this gap to a geometric separation in the shared embedding space, 
where visual and textual embeddings remain distinctly apart in the feature space.
This modality gap is further exacerbated in CL, as task-isolated training prevents joint calibration, reducing the text-based classifier's ability to adapt to new classes.

To mitigate semantic interference in the backbone, we consider it essential to prioritize the transfer of semantically relevant historical knowledge.
Since the text branch provides consistent semantics throughout continual adaptation, and CLIP inherently offers an aligned visual-textual space, we assume that \emph{textual cues offer reliable guidance for determining which past visual knowledge to transfer and which to suppress}, as shown in Fig.\ref{fig: intro}(b).
Moreover, to overcome the modality gap during classification, a straightforward solution lies in hybrid classifiers that support both cross-modality and intra-modality matching.
While visual classifiers based on raw prototypes often suffer from semantic inaccuracies due to limited data and class imbalance in real-world CL, 
we propose to inject the stable and structured semantics from the textual branch into the visual side, 
enabling a semantic-enriched visual classifier to bridge the modality gap.

Building on the above analysis, 
we propose Semantic-Enriched Continual Adaptation (SECA), a unified framework that harnesses the anti-forgetting and structured nature of textual semantic priors in CLIP. 
Specifically, we propose a Semantic-Guided Adaptive Knowledge Transfer (SG-AKT) module for selective transfer in the backbone.
It exploit the textual embeddings of new images as semantic vectors to assess their relevance to a pool of adapters that encapsulate historical visual knowledge. 
Guided by the relevance scores, past representations from the adapter pool are aggregated in an instance-adaptive manner and serve as teacher signals for selective distillation.
Besides, a Semantic-Enhanced Visual Prototype Refinement (SE-VPR) module is introduced to enable a powerful visual-side classifier.
It models the inter-class semantic relationships encoded in class-wise textual embeddings and uses them to refine the structure of coarse CLIP visual prototypes, aligning them with the relational topology of the textual space. 
Combined with the text classifier, it forms a hybrid classification paradigm that compensates for the modality gap and enhances plasticity.

Our main contributions are summarized as follows:

\begin{itemize}

    \item  
     We propose Semantic-Guided Adaptive Knowledge Transfer (SG-AKT), a novel instance-adaptive distillation approach that uses textual semantics to guide the selective transfer of historical knowledge, mitigating knowledge interference in the visual backbone;
         
    \item 
    We propose Semantic-Enhanced Visual Prototype Refinement (SE-VPR), which injects inter-class textual semantics into visual prototypes to build a powerful visual-side classifier, effectively compensating for the modality gap;
    
    \item 
    Extensive experiments demonstrate that our SECL outperforms existing methods and provides new insights into harnessing the potential of textual priors for CL.
        
\end{itemize}
\section{Related Work}

\begin{figure*}[!t]
\centering
\includegraphics[width=1.0\textwidth]{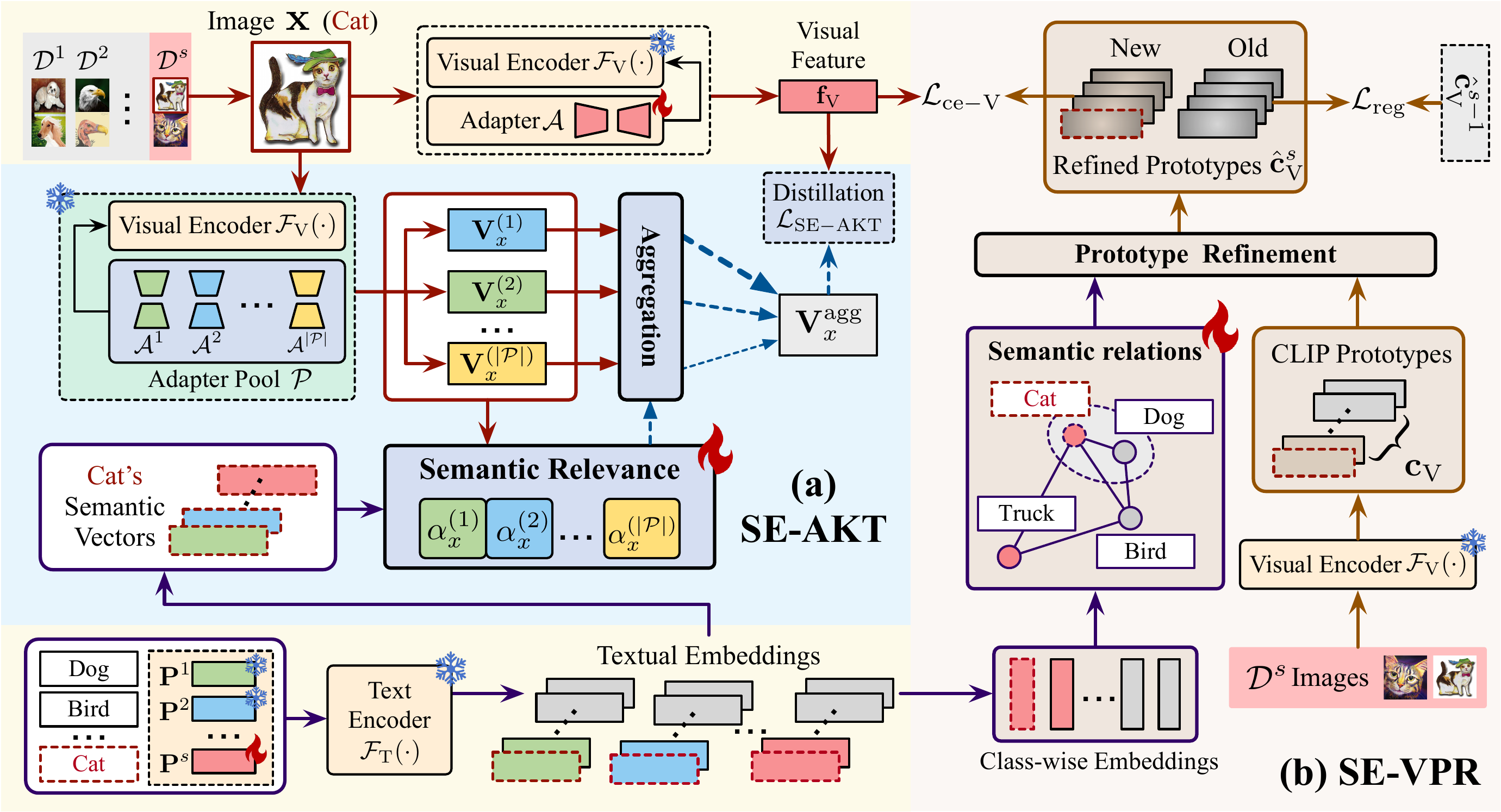}
\vspace{-4.5mm}
\caption{
{
Overall framework.
Our SECA is composed of two novel components: 
(a) 
The SG-AKT module that utilizes textual semantic vectors to aggregates relevant visual representations from a pool of historical adapters for distillation;
(b) 
The SE-VPR module that
leverages inter-class semantic relationships to refine CLIP prototypes, constructing a powerful visual-side classifier.
}
}\label{fig: framework}
\vspace{-5.0mm}
\end{figure*}

\textbf{Continual Learning.} 
Continual Learning (CL) aims to incrementally acquire new knowledge while preserving performance on previously learned tasks.
Traditional CL methods typically learn models from scratch and can be categorized into three types:
(a) replay-based methods \cite{replay1, replay2, replay3}, which store exemplar samples to jointly train with new tasks;
(b) regularization-based methods \cite{regularization1, regularization2, regularization3, wanghuaijie2025ekpc}, which constrain parameter updates through additional regularization terms; and
(c) expansion-based methods \cite{expansion1, expansion2}, which dynamically allocate new parameters for future tasks to improve model plasticity.
With the advent of large-scale pre-trained models \cite{ViT, 2021clip}, recent studies \cite{l2pCVPR22, vptnspNIPS-24, infloraCVPR24, helingfeng2025ckaa} have explored Parameter-Efficient Fine-Tuning (PEFT) as a promising solution for CL, enabling task adaptation without full model retraining.
In particular, CLIP-based continual learning \cite{rapfECCV24, PROOF-TPAMI25, clap4clip-NIPS24, menabue2024semantic} has attracted increasing interest due to its strong zero-shot generalization.
STAR~\cite{menabue2024semantic} treats the pre-trained CLIP as a powerful task-id selector.
RAPF \cite{rapfECCV24} introduces a lightweight learnable projector and a selective pseudo-feature replay approach, enhances fine-grained class discrimination.
PROOF \cite{PROOF-TPAMI25} improves image-text alignment via a learnable cross-attention module and alleviates forgetting through a weight averaging strategy.

\textbf{Parameter-Efficient Fine-Tuning (PEFT) for Vision-Lanugage Models.}
Recent vision-language models (VLMs) such as CLIP~\cite{2021clip} and BLIP~\cite{BLIP} enable strong zero-shot transfer but suffer from domain shift on downstream tasks. To address this, PEFT methods adapt VLMs by updating only a small set of parameters.
Prompt tuning-based methods ($e.g.$, CoOp~\cite{PromptTuning}, CoCoOp~\cite{cocoop}, ProDA~\cite{proda}, KgCoOp~\cite{KG-CoOp}) learn soft prompts to condition the text branch, while adapter-based approaches (e.g., CLIP-Adapter~\cite{CLIP-adapter}, Tip-Adapter~\cite{TIP-adapter}) introduce lightweight projection matrices or cache-based classifiers to improve model performance on downstream tasks.
Other PEFT strategies include low-rank tuning (LoRA~\cite{lora}), prefix tuning~\cite{PrefixTuning}, and Mixture-of-Experts (MoE~\cite{Mixture-of-Experts}).
Despite tuning only a small subset of parameters, these methods have demonstrated effectiveness comparable to full-model fine-tuning.

\section{Methodology}

\noindent \textbf{Problem Definition. } 
In continual learning (CL), the model is trained sequentially on a series of tasks $\mathcal{D} = \{\mathcal{D}^1, \cdots, \mathcal{D}^s, \cdots, \mathcal{D}^S \}$, where the task $\mathcal{D}^s = \{ (\mathbf{x}_i^s, y_i^s) \}_{i=1}^{\vert \mathcal{D}^s \vert}$ in $s$-th stage consists of image-label pairs from a distinct distribution.
The label set are non-overlapping, $i.e.$, $\mathcal{Y}^s \cap \mathcal{Y}^{s'} = \varnothing $ for any $s \neq s'$, where $\mathcal{Y}^s$ denotes the label space of $\mathcal{D}^s$. 
The goal is to incrementally learn all $S$ tasks while maintaining model performance on images from all seen classes $\mathcal{Y}^1 \cup  \cdots \cup \mathcal{Y}^S$ during inference.

\textbf{Overall framework.}
The overall framework of SECA is illustrated in Fig.\ref{fig: framework}. 
Our SECA is built upon CLIP, which includes a visual encoder $\mathcal{F}_{\mathrm{V}}(\cdot)$ and a text encoder $\mathcal{F}_{\mathrm{T}}(\cdot)$.
For efficient adaptation, we inserts learnable textual prompts and visual adapters into both CLIP’s visual and text encoders.
The proposed SECA mainly consists of two components:
(1) Semantic-Guided Adaptive Knowledge Transfer (SG-AKT, Fig.~\ref{fig: framework}(a)), an instance-adaptive distillation approach that utilizes textual cues to estimate semantic relevance and selectively aggregate past knowledge. 
The aggregated representations act as teacher signals for distillation.
(2) Semantic-Enhanced Visual Prototype Refinement (SE-VPR, Fig.~\ref{fig: framework}(b)), a module that leverages inter-class semantic relations to refine the structure of visual prototypes, which are then used as a visual-side classifier to support hybrid classification.





\subsection{Hybrid Parameter-Efficient Fine-Tuning Baseline}\label{sec: hybrid_baselind}

\textbf{CLIP Fine-Tuning with Learnable Textual Prompts.} 
The CLIP architecture comprises a visual encoder $\mathcal{F}_\mathrm{V}(\cdot)$ and a text encoder $\mathcal{F}_\mathrm{T}(\cdot)$.
For the $s$-th task $\mathcal{D}^{s}$, we introduce a learnable task-specific textual prompt $\mathbf{P}^s \in \mathbb{R}^{M \times d_\mathrm{T}}$, where $M$ is the number of prompt tokens and $d_\mathrm{T}$ is the embedding dimension of the text encoder.
Given a class label $y$, the input is a concatenation of the learnable prompt and its tokenized class name $\mathrm{CLASS}_{y}$: 

\begin{equation}
\mathbf{E}_{y}^s = [\mathbf{P}^s]_1[\mathbf{P}^s]_2 \cdots [\mathbf{P}^s]_M [\mathrm{CLASS}_{y}],
\end{equation}

\noindent where
$[\mathbf{P}^s]_m$ is the $m$-th learnable token in $\mathbf{P}^s$. 
This sequence is then fed into the frozen text encoder $\mathcal{F}_\mathrm{T}(\cdot)$ to generate the textual feature for $y$: $\mathcal{F}_\mathrm{T}(y; \mathbf{P}^s) = \mathcal{F}_\mathrm{T}(\mathbf{E}_y^s) \in \mathbb{R}^{d_{\mathrm{T}}}$.


\textbf{CLIP Fine-Tuning with Visual Adapters.} 
To adapt the visual encoder $\mathcal{F}_\mathrm{V}(\cdot)$ to downstream tasks, we introduce a set of task-shared lightweight adapters \cite{adapter} into the visual backbone.
At the $s$-th task, we denote the adapter set as $\mathcal{A} = \{ \mathcal{A}_{l} \}_{l=1}^{L}$, where each $\mathcal{A}_{l}$ is inserted alongside the $l$-th transformer block.
Given an input $\hat{\mathbf{h}}_l$ of the Feed-Forward Network (FFN) at layer $l$, the output is computed as:

\begin{equation}
    \quad
    \mathbf{h}_{l+1} = \hat{\mathbf{h}}_l + \mathrm{FFN}(\hat{\mathbf{h}}_l) + \mathcal{A}_{l}(\hat{\mathbf{h}}_l).
\end{equation}


\noindent Given an image $\mathbf{x} \in \mathcal{D}^s$, its final visual feature  is denoted as $\mathcal{F}_\mathrm{V}(\mathbf{x}; \mathcal{A}) \in \mathbb{R}^{d_\mathrm{V}}$, where $d_\mathrm{V}$ is the dimension of the representation from visual encoder. 


\textbf{Hybrid Fine-Tuning Baseline.}
The probability of $\mathbf{x}$ being classified to class $y$ is computed as:

\begin{equation}\label{eq: loss_base}
p^s (y \mid \mathbf{x}) = 
\frac{
\exp( \langle \mathcal{F}_\mathrm{V}(\mathbf{x}; \mathcal{A}) , \mathcal{F}_\mathrm{T}({y}; \mathbf{P}^s) \rangle  / \tau)}
{\sum\limits_{ y \in \mathcal{Y}^{s} } \exp( \langle \mathcal{F}_\mathrm{V}(\mathbf{x}; \mathcal{A}) , \mathcal{F}_\mathrm{T}(y; \mathbf{P}^s) \rangle  / \tau)},
\end{equation}

\noindent where $\tau$ is a temperature factor and $\langle \cdot, \cdot \rangle$ denotes the cosine similarity between $\ell_2$-normalized visual and textual features.
Given a data point $(\mathbf{x}, y) \in \mathcal{D}^s$, the learnable PEFT modules are trained through a standard cross-entropy loss:

\begin{equation}
    \mathcal{L}_{\mathrm{ce-T}} = - y \log p^s (y \mid \mathbf{x}),
\end{equation}

\noindent 
This formulation keeps the pre-trained CLIP frozen while enabling efficient representation adaptation in both visual and textual branches.


\subsection{Semantic-Guided Adaptive Knowledge Transfer}\label{sec: TA-VKT}


To enable selective knowledge transfer, we begin by extracting semantic vectors for new images, which are then used to assess their relevance to past representations from a pool of historical adapters.
The resulting relevance scores guide the instance-adaptive aggregation of past representations,
and the aggregated features serve as teacher signals for distillation.


\textbf{Knowledge Aggregation and Distillation.}
To continually record learned visual knowledge, we maintain a pool of historical adapters, which serves as a repository of knowledge from past tasks.
At the end of each task, the trained adapter is cached in a adapter pool $\mathcal{P} = \{ \mathcal{A}^1, \mathcal{A}^2, \cdots, \mathcal{A}^{\vert \mathcal{P} \vert} \}$ with a fixed size $\vert \mathcal{P} \vert$. 
To maintain a constant pool size, we monitor adapter utility score and prune the past adapter with the highest utility when incorporating a new one.

When a new task arrives ($i.e.$, $\mathcal{D}^s$), given a data point $(\mathbf{x}, y) \in \mathcal{D}^s$, we extract the visual knowledge $\mathbf{V}_x$ of $\mathbf{x}$ from the adapter pool $\mathcal{P}$ by encoding $\mathbf{x}$ through all adapters in $\mathcal{P}$:

\vspace{-3mm}
\begin{equation}
\mathbf{V}_x^{(p)} = \mathcal{F}_\mathrm{V}(\mathbf{x}; \mathcal{A}^p), \quad
\mathbf{V}_x = [ \mathbf{V}_x^{(1)}, \cdots, \mathbf{V}_x^{(\vert P \vert)} ],
\end{equation}

\noindent where $\mathbf{V}_x^{(p)} \in \mathbb{R}^{d_\mathrm{V}}$ is the visual representation of $\mathbf{x}$ encoded by the $p$-th adapter $\mathcal{A}^p \in \mathcal{P}$.
$\mathbf{V}_x$ collects diverse visual representations of $\mathbf{x}$ from different previous models.

In parallel, we collect the textual semantic vectors of class label $y$ associated with image $\mathbf{x}$.
To be specific, 
we combine the ground-truth word embedding $\mathrm{CLASS}_y$ with textual prompts from the new and all previous tasks:

\vspace{-2mm}
\begin{equation}
\mathbf{S}^{(p)}_{y} = \mathcal{F}_\mathrm{T}(y; \mathbf{P}^p),
\quad
\mathbf{S}_y = [ \mathbf{S}_y^{(1)}, \cdots, \mathbf{S}_y^{(s)} ],  
\end{equation}
\vspace{-2mm}

\noindent where $\mathbf{S}^{(p)}_{y} \in d_\mathrm{T}$ is the text feature obtained by concatenating the word embedding $\mathrm{CLASS}_y$ with the $p$-th task's prompt $\mathbf{P}^p$.
Each textual feature provides a task-specific semantic perspective of class $y$.
This scheme extracts comprehensive and abundant semantic embeddings of class $y$.


During training, consider the 
image $\mathbf{x}$, we assess its semantic relevance to representations from the adapter pool using the associated semantic vectors $\mathbf{S}_y$. 
To be specific, we introduce two learnable semantic projectors, $\mathbf{W}_{\mathrm{S}} \in \mathbb{R}^{d_\mathrm{T} \times d_\mathrm{V}}$ and $\mathbf{W}_{\mathrm{V}} \in \mathbb{R}^{d_\mathrm{V} \times d_\mathrm{V}}$, to map both $\mathbf{S}_y$ and $\mathbf{V}_x$ into a shared semantic space.
Afterwards, the relevance score between $\mathbf{x}$ and adapter $\mathcal{A}^p$ is formulated as follows:

\vspace{-2.0mm}
\begin{equation}\label{eq:ta-vkt-attention}
    \alpha_x^{(p)} =  
    \frac{1}{s} 
    \sum_{i=1}^{s}
    [\phi(\mathbf{S}^{(i)}_{y}) \mathbf{W}_{\mathrm{S}}]^{\top}
    [\phi(\mathbf{V}_x^{(p)}) \mathbf{W}_{\mathrm{V}}],
\end{equation}
\vspace{-1.0mm}

\noindent where $\phi(\cdot)$ denotes the LayerNorm \cite{Layernorm} operation to
to stabilize training.
$\alpha_x^{(p)}$ represents the relevance of its ground-truth class $y$ to learned visual knowledge from $p$-th adapter in shared semantic space.
The relevance scores then used to guide the aggregation of visual knowledge $\mathbf{V}^{(x)}$ from the adapter pool:

\begin{equation}
    \mathbf{V}^{\mathrm{agg}}_x = 
    \sum_{p=1}^{\vert \mathcal{P} \vert}
    \frac
    {
    \exp(\lambda \alpha_x^{(p)})
    }
    {
    \sum_{i=1}^{\vert \mathcal{P} \vert}\exp(\lambda \alpha_x^{(i)})
    } \mathbf{V}^{(p)}_x,
\end{equation}

\noindent where the scaling factor $\lambda$ is introduced to promote smoother feature aggregation.
Such an aggregation paradigm adaptively assigns higher weights to knowledge from past adapters that are closely matched to the textual semantic vectors for each instance, prioritizing aggregating relevant knowledge.


Two semantic projectors $\mathbf{W}_{\mathrm{S}}$ and $\mathbf{W}_{\mathrm{V}}$ are optimized by a cross-entropy loss $\mathcal{L}_{\mathrm{agg}}$, which guarantee the alignment between visual representations $\mathbf{V}_x$ and textual vectors $\mathbf{S}_y$:

\begin{equation}
    \mathcal{L}_{\mathrm{agg}} = 
    -y \log p^s (y \mid \mathbf{V}^{\mathrm{agg}}_x).
\end{equation}

The aggregated representation $\mathbf{V}^{\mathrm{agg}}_x$ serves as a teacher representation to distill relevant knowledge into the current model.
The distillation is implemented by minimizing the KL divergence between the predictions of the aggregated feature $\mathbf{V}^{\mathrm{agg}}_x$ and feature $\mathbf{f}_{\mathrm{V}}=\mathcal{F}_{\mathrm{V}}(\mathbf{x}, \mathcal{A})$ from the current adapter:

\begin{equation}
\begin{aligned}
& \mathcal{L}_{\mathrm{SG\text{-}AKT}} = \\
& \sum\limits_{y \in \mathcal{Y}^s} 
\ \text{sg} \left( p^s(y \mid \mathbf{V}^{\mathrm{agg}}_x, \tau') \right)
\log 
\frac{
    \text{sg} \left( p^s(y \mid \mathbf{V}^{\mathrm{agg}}_x, \tau') \right)
}{
    p^s(y \mid \mathbf{f}_{\mathrm{V}}, \tau') + \epsilon
}.
\end{aligned}
\end{equation}



\noindent 
where $\mathrm{sg}(\cdot)$ denotes the stop-gradient operation, $\tau'$ is a temperature factor used to soften the probability distribution, similar to the role of $\tau$ in Eq.~\ref{eq: loss_base}, and $\epsilon$ is a small constant added for numerical stability.





\textbf{Adapter Pool Management via Utility Score.}
To effectively manage the fixed capacity of the adapter pool, 
we adopt a pruning strategy: when incorporating a new adapter, we remove the existing one with the highest utility score.
We define the utility score of the $p$-th adapter as $\mathrm{U}^p$, 
which represents the adapter's accumulated relevance to the current task.
It is initially set to a uniform value of $\frac{1}{\vert \mathcal{P} \vert}$.
During the aggregation process, the utility scores are updated using a momentum-based strategy based on $\alpha_x^{(p)}$:

\vspace{-2.0mm}
\begin{equation}
    \mathrm{U}^{p} \leftarrow \mu \mathrm{U}^{p} + (1 - \mu) \alpha^{(p)}_x, \quad p \in \{1, \cdots, \vert \mathcal{P} \vert\}, 
\end{equation}
\vspace{-2.0mm}

\noindent where $\mu$ is a momentum factor.
After training, a higher utility score indicates the adapter's knowledge has been sufficiently transferred to the latest model, making it a suitable candidate for removal to prune redundant knowledge while preventing the pool size from growing linearly with the number of tasks.


\subsection{Semantic-Enhanced Visual Prototype Refinement}\label{sec: TG-VPR}

To bridge the modality gap, we construct a semantic-guided visual-side classifier for collaborative inference with the textual classifier.
This is achieved by modeling inter-class semantic relationships via textual embeddings, which are then used to refine the structure of coarse visual prototypes, aligning them with semantic structures in the text branch.



\textbf{Prototype Refinement.}
Consider the $s$-th task,
for each seen class within and before the $s$-th task (with class label set $\mathcal{Y}^{1:s} = \mathcal{Y}^1 \cup \cdots \cup \mathcal{Y}^s$),
we obtain a semantic embedding:
Given a class $k$ and the task-specific prompt $\mathbf{P}^s$, we encode the class name using CLIP’s text encoder: $\mathbf{Z}_k = \mathcal{F}_\mathrm{T}(k; \mathbf{P}^s)$.

To enable flexible modeling of inter-class relationships, we introduce a learnable projector $\mathbf{H}_{\mathrm{proj}} \in \mathbb{R}^{d_\mathrm{T} \times d_\mathrm{T}}$ to project the class-wise embeddings into a trainable and more expressive latent space.
We model the inter-class affinity matrix $\mathbf{M}$, which captures semantic similarities between different classes, based on the projected class-wise embeddings:


\begin{equation}
    \mathbf{M}_{k,j} = \exp(-\gamma \Vert \phi(\mathbf{Z}_k)\mathbf{H}_{\mathrm{proj}} -\phi(\mathbf{Z}_j)\mathbf{H}_{\mathrm{proj}} \Vert_2^2),
\end{equation}

\noindent where $\mathbf{M}_{k,j}$ represents the affinity score between the $k$-th and $j$-th classes, and $\gamma$ is a scaling factor to smooth the distribution of affinity values.
Subsequently, the affinity scores are used to refine the structure of coarse visual prototypes:

\begin{equation}
    \hat{\mathbf{c}}_{\mathrm{V},k} = 
    \sum_{j \in \mathcal{Y}^{1:s}}
    \frac
    {
    \mathbf{M}_{k,j}
    }
    {
    \sum_{i \in \mathcal{Y}^{1:s}} \mathbf{M}_{k,i}
    } {\mathbf{c}}_{\mathrm{V},k},
\end{equation}

\noindent
where ${\mathbf{c}}_{\mathrm{V},k}$ denotes the coarse CLIP prototype of the $k$-th class.
It is calculated by averaging the CLIP visual encoder features of all images belonging to class $k$ at the beginning of its corresponding task:

\begin{equation}
    \mathbf{c}_{\mathrm{V}, k} = 
    \frac{1}{N_k}
    \sum_{(\mathbf{x}, y) \in \mathcal{D}^{1:s}} \mathcal{F}_\mathrm{V}(\mathbf{x}) \mathds{1}(y=k),
\end{equation}

\noindent
where $N_k$ is the number of images of class $k$. 
$\mathds{1}(y=k)$ is an indicator function that equals 1 if $y=k$, and 0 otherwise.

The refined visual prototypes $\hat{\mathbf{c}}_\mathrm{V}^s = [\hat{\mathbf{c}}_{\mathrm{V}, 1}^s, \cdots, \hat{\mathbf{c}}_{\mathrm{V}, \vert \mathcal{Y}^{1:s} \vert }^s]$ serve as a visual-side classifier.
For $(\mathbf{x}, y) \in \mathcal{D}^s$,
the probability of predicting $\mathbf{x}$ as class $y \in \mathcal{Y}^s$ is computed as:

\vspace{-1mm}
\begin{equation}
    \hat{p}_\mathrm{V}(y \mid \mathbf{x}) = 
    \frac{
    \exp( \langle \mathcal{F}_\mathrm{V}(\mathbf{x}; \mathcal{A}) , \hat{\mathbf{c}}^s_{\mathrm{V}, y} \rangle  / \tau)}
    {\sum\limits_{ y \in \mathcal{Y}^{s}} \exp( \langle \mathcal{F}_\mathrm{V}(\mathbf{x}; \mathcal{A}) , \hat{\mathbf{c}}^s_{\mathrm{V}, y} \rangle  / \tau)}.
\end{equation}
\vspace{0.5mm}

The projector $\mathbf{H}_{\mathrm{proj}}$ is trained with a classification loss $\mathcal{L}_{\mathrm{ce-V}}$ to instruct the inter-class relationship modeling:

\begin{equation}
\mathcal{L}_{\mathrm{ce-V}} = - y \log \hat{p}_\mathrm{V}(y \mid \mathbf{x}). 
\end{equation}
\vspace{0.5mm}

\textbf{Prototype Consistency Regularization.}
Furthermore, to prevent $\mathbf{H}_{\mathrm{proj}}$ from overfitting to new classes, we introduce a prototype regularization loss $\mathcal{L}_{\mathrm{reg}}$ that promotes the temporal stability of visual prototypes for previously learned classes ($\mathcal{Y}^{1:s-1}$).  
This loss encourages the updated prototypes associated with old tasks to remain close to their earlier versions:


\begin{equation}
\mathcal{L}_{\mathrm{reg}} = \frac{1}{\vert \mathcal{Y}^{1:s-1} \vert}
\sum_{k \in \mathcal{Y}^{1:s-1} }
\Vert \hat{\mathbf{c}}_{\mathrm{V}, k}^s - \hat{\mathbf{c}}_{\mathrm{V}, k}^{s-1} \Vert^2_2,
\end{equation}

\noindent 
where $\hat{\mathbf{c}}_{\mathrm{V}, k}^{s-1}$ is the refined visual prototype after the $(s-1)$-th task of the previous class $k$.
By injecting textual semantics into visual prototypes and preserving temporal consistency, SE-VPR establishes a powerful visual-side classifier to compensate for the modality gap.

\subsection{Training Objective and Inference Strategy}

\textbf{Training Objective.} Our framework optimizes the SG-AKT and the SE-VPR modules, including the following terms:

\begin{equation}
\mathcal{L} = 
\mathcal{L}_{\mathrm{ce-T}} +
\underbrace{ \mathcal{L}_{\mathrm{agg}} +
\beta \mathcal{L}_{\mathrm{SG-AKT}} }_{\mathrm{SG-AKT}} +
\underbrace{ \mathcal{L}_{\mathrm{ce-V}} + 
\mathcal{L}_{\mathrm{reg}} }_{\mathrm{SE-VPR}},
\end{equation}

\noindent
where $\beta$ is a trade-off hyper-parameter controlling the contribution of $\mathcal{L}_{\mathrm{SG-AKT}}$.
Due to the growing accumulation of old task knowledge during continual adaptation, we set $\beta$ as a task-dependent variable that increases with tasks.

\textbf{Inference Strategy.} 
During inference, after the $s$-th task, given an image $\mathbf{x}$, its final prediction $\hat{y}$ is in a hybrid form, combining the prediction $\hat{p}_\mathrm{V}$ of the refined visual classifier $\hat{\mathbf{c}}_\mathrm{V}$ and predictions from all task-specific text classifiers:


\begin{equation}
\hat{y} = 
\arg\max_{y} \left(
\hat{p}_\mathrm{V}(y \mid \mathbf{x}, \tau') + 
\frac{1}{s} \sum_{i=1}^s p^i (y \mid \mathbf{x}, \tau')
\right),
\end{equation}

\noindent
where $\tau'$ is the temperature factor for inference, which is set to match the one used in $\mathcal{L}_{\mathrm{SG-AKT}}$ to ensure consistency between training and inference.

\section{Experiments}\label{sec: experiments}

\begin{table*}[t]
\centering
\caption{{Experimental results on ImageNetR and ImageNetA. We report the averaged results over 3 trials. 
VPT-NSP$^\dag$ and RAPF$^\dag$ denote the feature replay-free version of VPT-NSP~\cite{vptnspNIPS-24} and RAPF~\cite{rapfECCV24}, respectively.
The highest results are in \textbf{bold}, and the second best results are \underline{underlined}.}}
\label{tab: comparison_imagenet1}
\vspace{-3mm}
\begin{adjustbox}{max width=1.0\textwidth}
\begin{tabular}{
lccccccccccc}
\toprule
\multirow{2}{*}{Method} &
\multirow{2}{*}{Venue} & 
\multirow{2}{*}{Backbone} &
\multirow{2}{*}{ER/FR} &
\multicolumn{2}{c}{10S-ImageNetR} & \multicolumn{2}{c}{10S-ImageNetA} & \multicolumn{2}{c}{20S-ImageNetR} & 
\multicolumn{2}{c}{20S-ImageNetA}
\\ \cmidrule(l){5-12} 	
& & & & Last \uar & Avg \uar & Last \uar & Avg \uar & Last \uar & Avg \uar & Last \uar & Avg \uar \\
\midrule
L2P \cite{l2pCVPR22} & CVPR-22 & ViT-21k & \textbf{\ding{55}}  & 72.34 & 77.36 & 44.04 & 51.24 & 69.64 & 75.28 & 40.48 & 49.62 \\
CODA \cite{codaCVPR23} & CVPR-23 & ViT-21k & \textbf{\ding{55}} & 73.31 & 78.47 & 52.08 & 63.92 & 69.96 & 75.34 & 44.62 & 54.86 \\
RanPAC \cite{ranpac-NIPS23} & NeurIPS-23 & ViT-21k & \textbf{\ding{55}} & 77.90 & 82.91 & 62.40 & 67.58 & -- & -- & -- & -- \\
Adam-adapter \cite{adamIJCV2024} & IJCV-24 & ViT-21k & \textbf{\ding{55}} & 65.79 & 72.42 & 48.81 & 58.84 & 57.42 & 64.75 & 48.65 & 59.55 \\
SSIAT \cite{ssaitCVPR24} & CVPR-24 & ViT-21k & \textbf{\ding{51}} & 79.38 & 83.63 & 62.43 & 70.83 & 75.67 & 82.30 & 59.16 & 68.45 \\
VPT-NSP \cite{vptnspNIPS-24} & NeurIPS-24 & ViT-21k & \textbf{\ding{55}} & 77.95 & 83.44 & 53.83 & 63.93 & 75.69 & 81.87 & 49.81 & 61.41 \\
DIA \cite{DIA-CVPR2025} & CVPR-25 & ViT-21k & \textbf{\ding{51}} & 79.03 & 85.61 & 59.78 & 70.43 & 76.32 & 83.51 & -- & -- \\
ACMap \cite{ACMap-CVPR2025} & CVPR-25 & ViT-21k & \textbf{\ding{55}} & 73.50 & 79.50 & 56.19 & 65.19 & -- & -- & -- & -- \\
\midrule
ZS-CLIP & -- & CLIP & \textbf{\ding{55}} & 74.93 & 81.56 & 47.33 & 58.35 & 74.93 & 82.09 & 47.33 & 59.36 \\
L2P \cite{l2pCVPR22} & CVPR-22 & CLIP & \textbf{\ding{55}} & 75.98 & 81.67 & 47.86 & 59.35 & 68.78 & 76.87 & 47.54 & 59.77 \\
DualPrompt \cite{dualpromptECCV22} & CVPR-22 & CLIP & \textbf{\ding{55}} & 75.77 & 82.01 & 48.18 & 59.05 & 69.41 & 77.07 & 48.05 & 60.22 \\
CODA \cite{codaCVPR23} & CVPR-22 & CLIP & \textbf{\ding{55}} & 67.52 & 78.00 & 50.24 & 64.32 & 64.53 & 75.23 & 49.95 & 65.08 \\
Adam-adapter \cite{adamIJCV2024} & IJCV-24 & CLIP & \textbf{\ding{55}} & 71.35 & 78.65 & 59.35 & 68.56 & 68.75 & 76.71 & 58.55 & 67.72 \\
RAPF$^{\dag}$ \cite{rapfECCV24} & ECCV-24 & CLIP & \textbf{\ding{55}} & 73.23 & 82.20 & 45.54 & 60.67 & 71.28 & 81.66 & 43.85 & 58.54 \\
RAPF \cite{rapfECCV24} & ECCV-24 & CLIP & \textbf{\ding{51}} & 79.62 & 86.28 & 55.37 & 67.32 & 80.28 & 85.58 & 49.85 & 65.28 \\
CLAP \cite{clap4clip-NIPS24} & NeurIPS-24 & CLIP & \textbf{\ding{51}} & 79.98 & 85.77 & 58.66 & 69.35 & 79.12 & 85.03 & 55.84 & 67.72 \\
VPT-NSP$^{\dag}$ \cite{vptnspNIPS-24} & NeurIPS-24 & CLIP & \textbf{\ding{55}} & 77.45 & 83.60 & 47.14 & 59.04 & 77.52 & 83.94 & 47.07 & 59.88 \\
VPT-NSP \cite{vptnspNIPS-24} & NeurIPS-24 & CLIP & \textbf{\ding{51}} & 82.48 & 87.94 & 61.42 & 71.76 & 82.06 & 88.09 & 60.70 & 72.57 \\
PROOF \cite{PROOF-TPAMI25} & T-PAMI-25 & CLIP & \textbf{\ding{51}} & 77.25 & 82.69 & 55.67 & 65.50 & 77.05 & 82.83 & 54.05 & 64.53 \\
\midrule
SECA (Ours) & -- & CLIP & \textbf{\ding{55}} & \underline{83.18}	& \underline{88.58} & \underline{65.09} & \underline{74.45} & \underline{82.25} & \underline{88.19} & \underline{62.80} & \underline{73.02} \\
SECA++ (Ours) & -- & CLIP & \textbf{\ding{51}} &  \textbf{83.41} & \textbf{88.75} & \textbf{65.77} & \textbf{74.65} & \textbf{83.02} & \textbf{88.62} & \textbf{64.41} & \textbf{74.64} \\
\bottomrule
\end{tabular}
\end{adjustbox}
\vspace{-2mm}
\end{table*}
\begin{table}[ht]
\centering
\caption{\footnotesize{Experimental results on CIFAR100. All results are based on the CLIP model with a ViT-B/16 vision backbone.}}\label{tab: comparison_cifar}
\vspace{-2mm}
\begin{adjustbox}{max width=0.5\textwidth}
\begin{tabular}{lccccc}
\toprule
\multirow{2}*{Method} & 
\multirow{2}*{ER/FR} &
\multicolumn{2}{c}{10S-CIFAR100} & \multicolumn{2}{c}{20S-CIFAR100} \\
\cmidrule{3-6}
& & Last \uar & Avg \uar & Last \uar &Avg \uar \\
\midrule
ZS-CLIP & \textbf{\ding{55}} & 66.68       & 75.15        & 66.68        & 75.93       \\
L2P \cite{l2pCVPR22} & \textbf{\ding{55}} & 73.08 & 81.90 & 68.67 & 79.18 \\
DualPrompt \cite{dualpromptECCV22} & \textbf{\ding{55}} & 72.51         & 81.45        & 69.91        & 79.74       \\
CODA \cite{codaCVPR23}     & \textbf{\ding{55}} & 62.25         & 76.98        & 41.98        & 69.78       \\
SLCA \cite{slcaICCV23}         & \textbf{\ding{51}} & 67.58         & 80.53        & 66.84        & 78.96       \\
Adam-adapter \cite{adamIJCV2024} & \textbf{\ding{55}} & 65.50  & 75.76 & 58.12  & 70.18 \\
RAPF \cite{rapfECCV24}              & \textbf{\ding{51}} & 79.04  & 86.19 & \underline{79.26} & \underline{86.87} \\ 
CLAP \cite{clap4clip-NIPS24} & \textbf{\ding{51}} & 78.21 & 86.13 & 77.35 & 86.08 \\
PROOF \cite{PROOF-TPAMI25} & \textbf{\ding{51}} & 76.29 & 84.88 & 76.13 & 85.12 \\
\midrule
SECA (Ours) & \textbf{\ding{55}} & \underline{79.79} & \underline{86.70} & 77.73 & 85.35 \\
SECA++ (Ours) & \textbf{\ding{51}} & \textbf{81.59} & \textbf{87.80} & \textbf{80.07} & \textbf{87.11} \\
\bottomrule
\end{tabular}
\end{adjustbox}
\vspace{-3mm}
\end{table}


\textbf{Datasets.} We evaluate our method on three representative CIL benchmarks: 
ImageNetR~\cite{imagenet-r}, ImageNetA~\cite{imagenet-a}, and CIFAR-100~\cite{cifar100}.
ImageNetR comprises 30,000 images spanning 200 classes, covering a wide range of visual styles.
ImageNetA contains 7,500 challenging real-world adversarial examples images across 200 classes.
CIFAR100 is a commonly-used dataset in continual learning, which consists of 60000 32$\times$32 images of 100 classes.
We adopt both 10-split and 20-split settings, where the class set is evenly divided into 10 or 20 non-overlapping tasks.

\noindent
\textbf{Evaluation Metrics.}
Following prior PEFT-based CIL works~\cite{ssaitCVPR24, cpromptCVPR24, rapfECCV24}, we adopt two standard evaluation metrics:
(1) The last session accuracy (Last): the final accuracy over all classes after completing the last task, and
(2) Average accuracy (Avg), the average accuracy across all incremental tasks.
All experiments are conducted using three random seeds, and we report the mean performances.

\noindent
\textbf{Implementation Details.} All experiments are conducted on a single NVIDIA RTX 3090 GPU. We adopt CLIP~\cite{2021clip} with a ViT-B/16~\cite{ViT} visual encoder pre-trained by OPENAI as our backbone.
Our model is optimized using Adam~\cite{adam_optimizer} with an initial learning rate of 0.001.
For the training schedule, each task is trained for 10 epochs on ImageNetR and ImageNetA, and 3 epochs on CIFAR100.
The batch size is set to 64 for ImageNetR and ImageNetA and 100 for CIFAR100.
The capacity of the frozen adapter pool $\mathcal{P}$ in SG-AKT is set to 5,
and the temperature factor $\tau'$ in $\mathcal{L}_{\mathrm{SG-AKT}}$ is set to 0.05.

\noindent
\textbf{SECA++: An Enhanced Variant of SECA.}
Following existing PEFT-based CIL methods~\cite{ssaitCVPR24, rapfECCV24, vptnspNIPS-24}, we further implement an enhanced variant, denoted as SECA++, which incorporates Gaussian sampling-based feature replay to alleviate inter-task confusion during the training of multi-modal classifier, particularly for semantically confusing or fine-grained classes.

\subsection{Comparison with the State-of-the-Arts}

We compare SECA with recent state-of-the-art PEFT-based approaches, including prompt-based methods~\cite{l2pCVPR22, vptnspNIPS-24, codaCVPR23}, adapter-based methods~\cite{DIA-CVPR2025, ssaitCVPR24, rapfECCV24},
the zero-shot CLIP (ZS-CLIP) and other representative techniques built upon either ViT-B/16 pretrained on ImageNet-21K~\cite{imagenet-21k} (ViT-21K) or CLIP-pretrained backbones.
Evaluations are conducted on ImageNetR, ImageNetA and CIFAR100, 
and the results are shown in Tab.\ref{tab:comparison_imagenet} and Tab.\ref{tab: comparison_cifar}.
`ER/FP' indicates the use of experience replay~\cite{ER} or distributional sampling-based feature replay.
Notably, most CLIP-based CIL methods incorporate such replay strategies into their frameworks.
For a fair comparison, we reproduce the results of two representative methods, RAPF~\cite{rapfECCV24} and VPT-NSP~\cite{vptnspNIPS-24}, using a CLIP backbone under a replay-free setting.
These variants are denoted as RAPF$^\dag$ and VPT-NSP$^\dag$.

The results demonstrate the superiority of our method.
Specifically, under the 10-split setting, our replay-free SECA surpasses the strongest VPT-NSP (the replay-based version) by 0.70\%, 3.67\% on ImageNet-R and ImageNet-A in terms of Last accuracy, respectively.
Our feature replay-enhanced version, SECA++, further boosts the performance, outperforming VPT-NSP by 0.93\% and 4.35\% in Last accuracy on 10S-ImageNetR and 10S-ImageNetA, respectively.

To sum up, our SECA has two main advantages:
(1) It enables seamless knowledge injection into each layer within the visual encoder via layer-wise adapters combined with a adaptive knowledge transfer mechanism, rather than merely attaching tunable modules after the frozen backbone;
(2) It is compatible with replay-based strategies, allowing further performance improvements when integrated with lightweight feature replay. 
\textbf{\emph{More evaluations can be found in Appendix~\ref{sec: experiments}.B.}}

\subsection{Ablation Studies}

\begin{table*}[!t]
\centering
\caption{{Ablation studies of each component in our approach on the three benchmarks.}}
\vspace{-2.0mm}
\label{tab:ablation}
\begin{adjustbox}{max width=1.0\textwidth}
\begin{tabular}{
ccccllllll}
\toprule
\multirow{2}{*}{Idx} &
\multirow{2}{*}{H-PEFT} 
& \multirow{2}{*}{SG-AKT} &
\multirow{2}{*}{SE-VPR} & \multicolumn{2}{c}{10S-ImageNetA} & \multicolumn{2}{c}{10S-CIFAR100} & \multicolumn{2}{c}{10S-ImageNetR}\\ 
\cmidrule(l){5-10} 	
& & & & Last \uar & Avg \uar & Last \uar & Avg \uar & Last \uar & Avg \uar \\
\midrule
1 & - & - & - & \raggedright 47.33 &58.35 & 67.19 & 76.23 & 74.93 & 81.56 \\
4 & \textbf{\ding{51}} & - & -  & 55.78\tsb{\(\pm\)1.36} & 67.97\tsb{\(\pm\)0.38} & 73.97\tsb{\(\pm\)0.65} & 82.75\tsb{\(\pm\)0.28} & 80.57\tsb{\(\pm\)0.15} & 87.04\tsb{\(\pm\)0.35}\\
5 & \textbf{\ding{51}} & \textbf{\ding{51}} & - & 57.91\tsb{\(\pm\)0.55} & 68.56\tsb{\(\pm\)0.66} & 75.93\tsb{\(\pm\)0.09} & 84.16\tsb{\(\pm\)0.21} & 81.36\tsb{\(\pm\)0.49} & 87.55\tsb{\(\pm\)0.29} \\
6 & \textbf{\ding{51}} & - & \textbf{\ding{51}}  
& 62.62\tsb{\(\pm\)0.65} & 73.15\tsb{\(\pm\)0.25} & 77.13\tsb{\(\pm\)0.56} & 84.91\tsb{\(\pm\)0.43} & 81.30\tsb{\(\pm\)0.55} & 87.63\tsb{\(\pm\)0.12}\\
7 & \textbf{\ding{51}} & \textbf{\ding{51}} & \textbf{\ding{51}} & \textbf{65.09}\tsb{\(\pm\)0.48} & \textbf{74.45}\tsb{\(\pm\)0.76} & \textbf{79.79}\tsb{\(\pm\)0.42} & \textbf{86.70}\tsb{\(\pm\)0.16} & \textbf{83.18}\tsb{\(\pm\)0.34} & \textbf{88.58}\tsb{\(\pm\)0.07} \\
\bottomrule
\end{tabular}
\end{adjustbox}
\vspace{-3.0mm}
\end{table*}

\textbf{The effectiveness of components.} 
We conduct ablation studies on components of our training pipeline, including the hybrid PEFT baseline (H-PEFT), SG-AKT and SE-VPR. 
The results are shown in Tab.\ref{tab:ablation}.
The \textbf{H-PEFT} (Idx 2) achieves a substantial improvement over ZS-CLIP (Idx 1), with +5.64\% and +8.45\% gains in Last accuracy on 10S-ImageNetR and 10S-ImageNetA.
Incorporating \textbf{SG-AKT} (Idx 3) brings further gains (e.g., +2.13\% on ImageNetA), effectively improving
representation learning in the visual encoder by selectively transferring past knowledge.
Moreover, \textbf{SE-VPR} (Idx 4) form a hybrid classification paradigm by leveraging textual semantics to refine visual prototypes, resulting in notable performance improvements (e.g., +4.65\% on ImageNetA) compared to text-only classifier.

\begin{table}[!t]
\renewcommand{\arraystretch}{0.9} 
\centering
\caption{{Ablation studies of different distillation strategies on 10S-ImageNetA and 10S-CIFAR100.}}\label{tab: ablation_KD}
\vspace{-1mm}
\begin{adjustbox}{max width=0.48\textwidth}
\begin{tabular}{lcccc}
\toprule
\multirow{2}*{Method} & 
\multicolumn{2}{c}{10S-ImageNetA} & \multicolumn{2}{c}{10S-CIFAR100} \\
\cmidrule{2-5}
& Last \uar & Avg \uar & Last \uar &Avg \uar \\
\midrule
\multicolumn{5}{l}{\textbf{Methods without SE-VPR.}} \\
\midrule
Seq. & 55.78\tsb{\(\pm\)1.36} & 67.97\tsb{\(\pm\)0.38} & 73.97\tsb{\(\pm\)0.65} & 82.75\tsb{\(\pm\)0.28} \\
CLIP-KD & 55.01\tsb{\(\pm\)0.68} & 67.00\tsb{\(\pm\)0.27} & 71.21\tsb{\(\pm\)0.61} & 80.98\tsb{\(\pm\)0.50} \\
Vanilla & 57.05\tsb{\(\pm\)0.75} & 67.93\tsb{\(\pm\)0.47} & 74.55\tsb{\(\pm\)0.26} & 83.26\tsb{\(\pm\)0.32} \\
Avg-KD & 56.90\tsb{\(\pm\)0.11} & 68.02\tsb{\(\pm\)0.76} & 75.05\tsb{\(\pm\)0.78} & 83.65\tsb{\(\pm\)0.33} \\
SG-AKT (Ours) & \textbf{57.91}\tsb{\(\pm\)0.55} & \textbf{68.56}\tsb{\(\pm\)0.66} & \textbf{75.93}\tsb{\(\pm\)0.09} & \textbf{84.16}\tsb{\(\pm\)0.21}
\\
\midrule
\multicolumn{5}{l}{\textbf{Methods with SE-VPR.}} \\
\midrule
Seq. & 62.62\tsb{\(\pm\)0.65} & 73.15\tsb{\(\pm\)0.25} & 77.13\tsb{\(\pm\)0.56} & 84.91\tsb{\(\pm\)0.43} \\
CLIP-KD & 62.39\tsb{\(\pm\)0.77} & 73.01\tsb{\(\pm\)1.17} & 73.30\tsb{\(\pm\)0.80} & 83.76\tsb{\(\pm\)0.18} \\
Vanilla & 63.51\tsb{\(\pm\)0.71} & 73.89\tsb{\(\pm\)0.69} & 76.42\tsb{\(\pm\)0.44} & 85.75\tsb{\(\pm\)0.37} \\
Avg-KD & 64.32\tsb{\(\pm\)0.59} & 74.08\tsb{\(\pm\)0.82} & 78.98\tsb{\(\pm\)0.51} & 86.47\tsb{\(\pm\)0.30} \\
SG-AKT (Ours) & \textbf{65.09}\tsb{\(\pm\)0.48} & \textbf{74.45}\tsb{\(\pm\)0.76} & \textbf{79.79}\tsb{\(\pm\)0.42} & \textbf{86.70}\tsb{\(\pm\)0.16} \\
\bottomrule
\end{tabular}
\end{adjustbox}
\vspace{-2.0mm}
\end{table}

\textbf{SG-AKT v.s. other Distillation Strategies.}
We compare our SG-AKT with four alternative training strategies in Tab.\ref{tab: ablation_KD}: 
(1) \textbf{Seq.} denotes sequential tuning without any distillation loss; 
(2) \textbf{CLIP-KD} distills from the frozen CLIP model as a global teacher; 
(3) \textbf{Vanilla} distills knowledge from the most recent task model, following prior works~\cite{AFC-CVPR22, PSRD-ICML23}; 
(4) \textbf{Avg-KD} is a simplified variant of SG-AKT that aggregates adapter pool features by averaging, without text-aware attention.
Tab.~\ref{tab: ablation_KD} shows that SG-AKT consistently outperforms all variants, with +0.77\% and +1.58\% Last-acc improvements over Avg-KD and Vanilla on 10S-ImageNetA, respectively.
Although these two designs are effective for knowledge retention, they fall short compared to SG-AKT, highlighting the effectiveness of leveraging textual information to prioritize semantically relevant transfer.


\textbf{SE-VPR v.s. other Classifier Designs.}
We compare our SE-VPR module with four alternative designs for constructing the classifier in Tab.\ref{tab: ablation_Prototype}:
(1) \textbf{Only Text} denotes using the text encoder alone for classification without visual classifiers;
(2) \textbf{Centroid (CLIP)} computes class centroids using pretrained frozen CLIP prototypes as the visual-side classifier;
(3) \textbf{Centroid (adapted)} leverages the fine-tuned CLIP visual encoder to extract class centroids;
(4) \textbf{Linear} trains an additional linear classifier on top of the adapted visual features.
The results indicate the necessity of incorporating a visual-side classifier to bridge the modality gap.
It also show the superiority of our SE-VPR, which achieves the best performance compared to other variants, yielding +1.93\% and +1.49\% gains in Last accuracy over Centroid (adapted) on 10S-ImageNetA and 10S-CIFAR100.

\begin{table}[!t]
\begin{minipage}[t]{0.47\textwidth}
\centering
\captionof{table}{{Ablation studies of different classifier designs on 10S-ImageNetA and 10S-CIFAR100 datasets.}}\label{tab: ablation_Prototype}
\vspace{-1mm}
\scalebox{0.80}{
\setlength{\tabcolsep}{2pt}  
\centering
\begin{tabular}{lcccc}
\toprule
\multirow{2}*{Method} & 
\multicolumn{2}{c}{10S-ImageNetA} & \multicolumn{2}{c}{10S-CIFAR100} \\
\cmidrule{2-5}
& Last \uar & Avg \uar & Last \uar &Avg \uar \\
\midrule
Only Text & 57.91\tsb{\(\pm\)0.55} & 68.56\tsb{\(\pm\)0.66} & 75.93\tsb{\(\pm\)0.09} & 84.16\tsb{\(\pm\)0.21} 
\\
Centroid (CLIP) & 58.55\tsb{\(\pm\)0.57} & 68.30\tsb{\(\pm\)1.17} & 75.77\tsb{\(\pm\)0.56} & 83.88\tsb{\(\pm\)0.42}
\\
Centroid (Adapted) & 63.16\tsb{\(\pm\)0.30} & 72.89\tsb{\(\pm\)0.98} & 78.30\tsb{\(\pm\)0.89} & 86.15\tsb{\(\pm\)0.66} \\
Linear & 51.07\tsb{\(\pm\)3.34} & 62.76\tsb{\(\pm\)3.19} & 74.20\tsb{\(\pm\)0.50} & 82.98\tsb{\(\pm\)0.59}
\\
SE-VPR (Ours) & \textbf{65.09}\tsb{\(\pm\)0.48} & \textbf{74.45}\tsb{\(\pm\)0.76} & \textbf{79.79}\tsb{\(\pm\)0.42} & \textbf{86.70}\tsb{\(\pm\)0.16} \\
\bottomrule
\end{tabular}}
\vspace{-5mm}
\end{minipage}
\end{table}

\subsection{Hyper-Parameter Analysis}

\begin{figure}[h]
\centering
\includegraphics[width=0.48\textwidth]{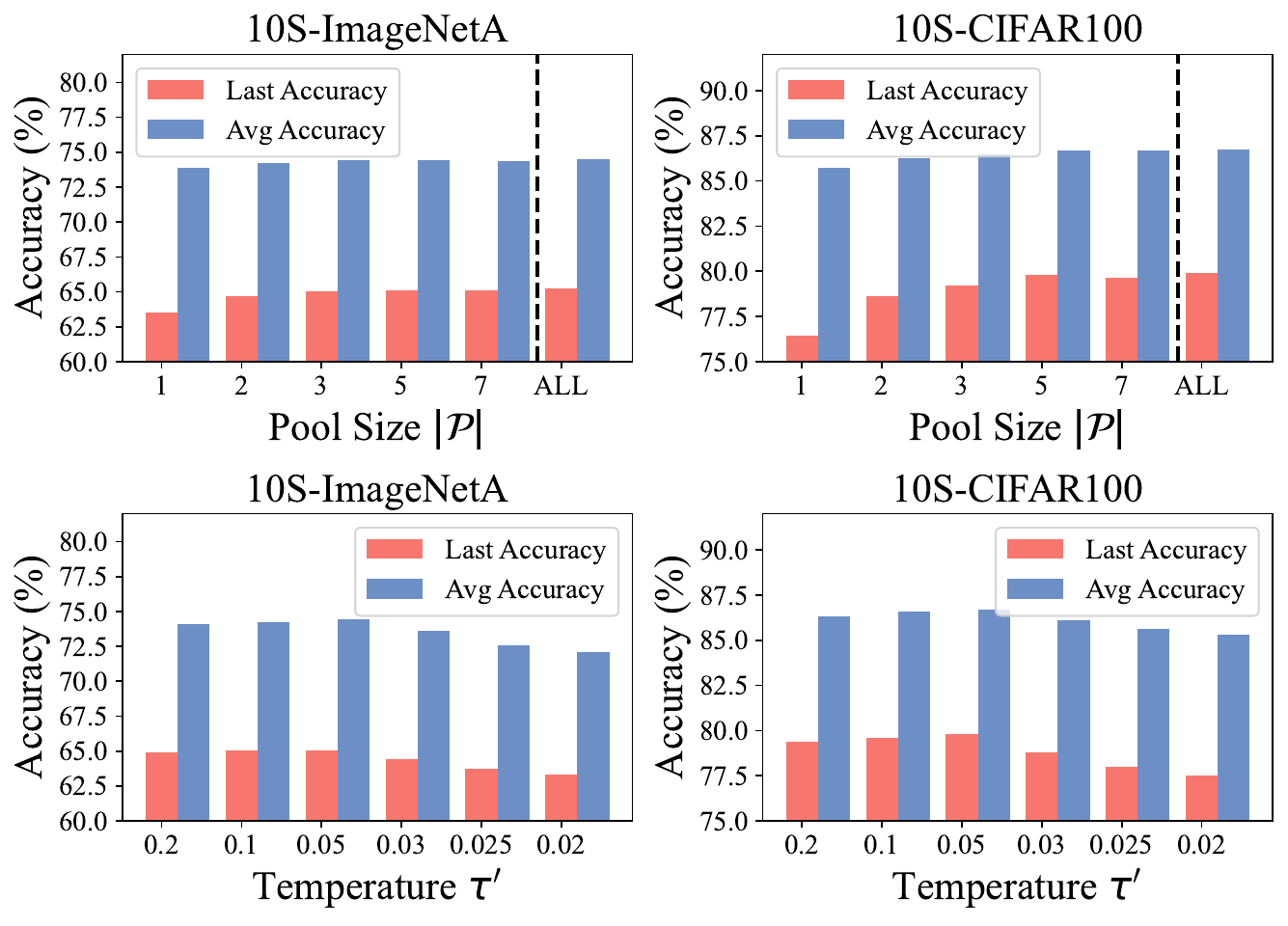}
\vspace{-6mm}
\caption{{Model performances under different pool sizes $\vert \mathcal{P} \vert$ and temperatures $\tau'$.}
}\label{fig: parameter}
\vspace{-8mm}
\end{figure}

\textbf{Analysis of the Pool Size $\vert \mathcal{P} \vert$ and the Temperature Factor $\tau'$.}
We analyze the sensitivity of two key hyper-parameters in SG-AKT: the size $\vert \mathcal{P} \vert$ of the adapter pool and the temperature factor $\tau'$ for knowledge transfer.
The results are shown in Fig.~\ref{fig: parameter}.
The performance constantly improves with larger pool size and saturates when $\vert \mathcal{P} \vert \ge 5$, where the results closely match those of using all previous adapters (denoted as `ALL').
This confirms that our utility-score-based pool management effectively prunes redundant knowledge without compromising accuracy. 
For the temperature factor $\tau'$, 
a relatively high $\tau'$ yields better performance, suggesting that smoother teacher labels benefits knowledge trransfer.
Based on the results, we set $\vert \mathcal{P} \vert = 5$ and $\tau' = 20.0$ for all datasets.

\section{Conclusion}

In this paper, we focus on the potential of textual semantics in addressing the stability–plasticity dilemma in continual learning (CL), and propose Semantic-Enriched Continual Adaptation (SECA), a unified framework for class-incremental learning with CLIP.
By assessing the new image's semantic relevance to historical knowledge using textual semantics,
our Semantic-Guided Adaptive Knowledge Transfer (SG-AKT) selectively transfers relevant past knowledge to the current model, 
suppressing the transfer of unrelated or conflicting knowledge.
Furthermore, by injecting inter-class semantic relations into visual prototypes, our Semantic-Enhanced Visual Prototype Refinement (SE-VPR) builds a powerful visual-side classifier to bridge the modality gap.
We believe SECA provides valuable insights into text-grounded continual learning by demonstrating how textual semantics enable better stability–plasticity trade-off.
In the future, we aim to further explore adaptive transfer and multi-modal synergy in CL.

\section{Acknowledgements}

This work was supported in part by the National Natural Science Foundation of China under Grants 62176198, 62576262, U22A2096, in part by Scientific and Technological Innovation Teams in Shaanxi Province under grant 2025RS-CXTD-011, in part by the Shaanxi Province Core Technology Research and Development Project under grant 2024QY2-GJHX-11, 2024GX-YBXM-135, in part by the Fundamental Research Funds for the Central Universities under Grant QTZX25083, QTZX23042. This work is also supported by the National Engineering Laboratory for Integrated Aero-Space-Ground- Ocean Big Data Application Technology, China

\bibliography{aaai25}

\appendix


\end{document}